\documentclass[10pt,twocolumn,letterpaper]{article}

\usepackage{cvpr}
\usepackage{times}
\usepackage{epsfig}
\usepackage{graphicx}
\usepackage{amsmath}
\usepackage{amssymb}
\usepackage{algorithm,algorithmic}
\usepackage{comment}

\usepackage[breaklinks=true,bookmarks=false]{hyperref}

\cvprfinalcopy 

\def\cvprPaperID{****} 

\begin{document}

\title{Convolutional Neural Networks with Transformed Input based on \\ Robust Tensor Network Decomposition}

\author{Jenn-Bing Ong, Wee-Keong Ng\\
Nanyang Technological University\\
{\tt\small ongj0063, awkng@e.ntu.edu.sg}
\and
C.-C. Jay Kuo\\
University of Southern California\\
{\tt\small cckuo@sipi.usc.edu}
}

\maketitle

\begin{abstract}
Tensor network decomposition, originated from quantum physics to model entangled many-particle quantum systems, turns out to be a promising mathematical technique to efficiently represent and process big data in parsimonious manner. In this study, we show that tensor networks can systematically partition structured data, e.g. color images, for distributed storage and communication in privacy-preserving manner. Leveraging the sea of big data and metadata privacy, empirical results show that neighboring subtensors with implicit information stored in tensor network formats cannot be identified for data reconstruction. This technique complements the existing encryption and randomization techniques which store explicit data representation at one place and highly susceptible to adversarial attacks such as side-channel attacks and de-anonymization. Furthermore, we propose a theory for adversarial examples that mislead convolutional neural networks to misclassification using subspace analysis based on singular value decomposition (SVD). The theory is extended to analyze higher-order tensors using tensor-train SVD (TT-SVD); it helps to explain the level of susceptibility of different datasets to adversarial attacks, the structural similarity of different adversarial attacks including global and localized attacks, and the efficacy of different adversarial defenses based on input transformation. An efficient and adaptive algorithm based on robust TT-SVD is then developed to detect  strong and static adversarial attacks.
\end{abstract}

\section{Introduction}
\begin{figure}[t]
\begin{center}
\includegraphics[width=1\linewidth]{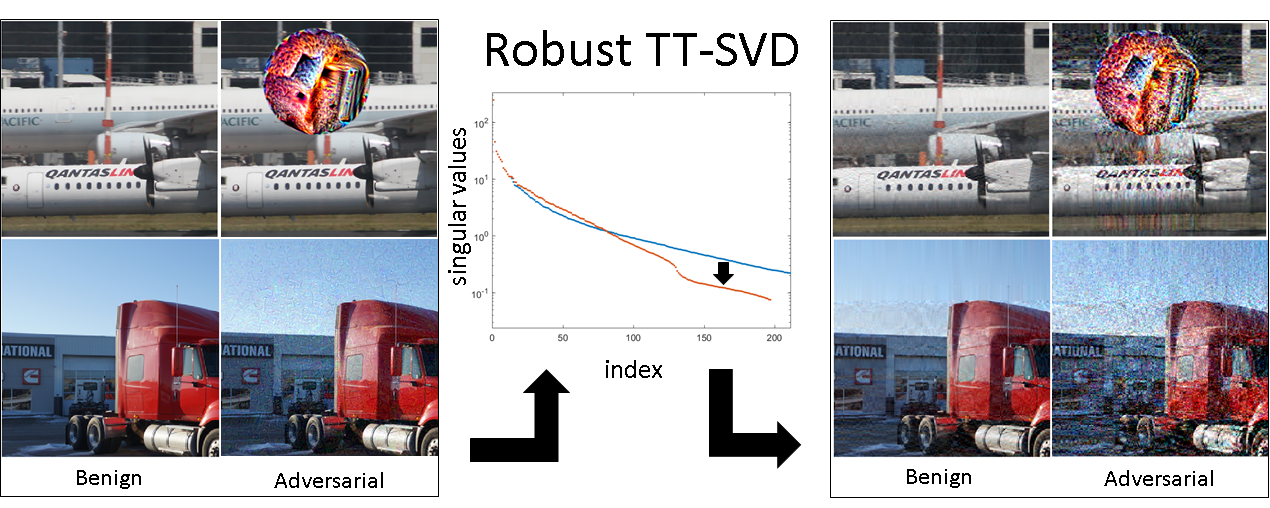}
\end{center}
   \caption{Global and localized adversarial examples, as diverse as their form can take, share similar structural properties in increasing the image roughness. This is because the sensitivity of subspace approximation by convolutional neural networks (CNNs) is controlled by the decay rate of singular values of the input image. The larger the decay rate, the smoother is the image input, and the more robust is the approximation. Our proposed robust TT-SVD algorithm linearly combines the singular values and vectors that fall within a (prescribed) bin to examine the robustness.}
\label{fig:key_points}
\end{figure}

Tensor computing recently emerges as a promising mathematical technique for big data processing and analytics~\cite{Cichocki2016,Cichocki2017}. Big data serves as the fuel in driving deep learning models that create tremendous value for various applications, ranging from science, business, to government. Deep learning automates the process of feature extraction and exploits their compositionality to construct high-level features that achieve human-level performance in many designated tasks such as classification and prediction~\cite{lecun2015}. The tradeoff, however, involves storage and processing of large amount of (labeled) data and models with millions to billions of parameters. The data explosion growth is expected to outpace the development of storage and processing technology, therefore domain-specific hardware acceleration and algorithmic codesign all aim to improve throughput and energy-efficiency without compromising model performance and hardware costs to cater for widespread deployment of deep learning models~\cite{Sze2017}. Sharing of personal and confidential data across organizations demand cutting-edge privacy-preserving technology. Current privacy-preserving technologies such as encryption and randomization techniques share a common drawback that any security breaches such as leakage of decryption key or the data content during the storage, communication, or computation phases expose the individual records that contain explicit information. Therefore, it is timely and essential to explore new data structures which provide not only efficient and distributed storage and computation, but also privacy preservation such that data leakage provides the adversary partial (implicit or latent) information of individual records and reconstruction is difficult without knowledge of the data structure.

Deep learning models, despite their impressive performance, are highly susceptible to adversarial attacks that attempt to perturb the inputs in subtle manner (imperceptible or quasi-imperceptible) to achieve the adversary's motives such as targeted or untargeted misclassifications. This has serious implications especially because these models are increasingly being deployed for mission-critical and safety-critical applications such as autonomous vehicles and robotics. Existing theories on adversarial examples such as models' linearity and non-linearity hypotheses, dimensional analyses, etc. do not generalize to different adversarial attacks, these theories typically stem from local empirical observations and do not fully align with each other~\cite{akhtar2018threat}. In this paper, we conduct both theoretical and experimental studies using tensor networks as data structure for the input data of convolutional neural networks (CNNs) and analyze their robustness to adversarial attacks. Our contributions include:
\begin{itemize}
\item Propose to use tensor network representations for distributed storage of input data for machine learning models. In particular, the model performance, storage, and compression / decompression efficiency are benchmarked for CNNs. Empirical results based on information theory show that neighboring subtensors with implicit / latent information cannot be identified for data reconstruction. The robustness of tensor network representations subject to perturbation of the subtensors is also investigated.
\item Propose a theory based on subspace analysis using singular value decomposition (SVD) for adversarial examples in CNNs. Using the theory, we quantify the robustness of different datasets to adversarial attacks, analyze the efficacy of different defense techniques based on input transformation and the structural similarity of different adversarial attacks. The theory is extended to analyze higher-order tensors with tensor-train SVD (TT-SVD) and an efficient algorithm is proposed to detect strong and static adversarial attacks (see Fig.~\ref{fig:key_points}).
\end{itemize}
The organization of this paper is as follows: Section~\ref{sec:tn} presents the preliminary knowledge of tensor networks such as tensor formats, properties, and storage complexity. The proposed robust TT-SVD algorithm for adversarial detection is presented in Section~\ref{subsec:robust_ttsvd}. Section 3 covers the threat model, adversarial attacks and defenses in CNNs. Experimental studies are conducted and the results are discussed in Section 4. Section 5 and 6 provide the related work and conclusion respectively.


\section{Tensor Networks (TN)}
\label{sec:tn}
Tensor decomposition has found many applications in signal processing and machine learning. Many review papers have been published throughout the years, more recent and relevant to machine learning and big data applications include~\cite{Cichocki2016,Cichocki2017,Sidiropoulos2016,Papalexakis2016,Cichocki2015,anandkumar2014,Grasedyck2013a}. The basic tensor formats and properties are summarized here.

\emph{Canonical Polyadic (CP) decomposition} is one of the most popular tensor technique due to the ease of interpretation. CP is expressed as the sum of rank-1 components
\begin{equation}
A(i_1,\cdots,i_d)\cong\sum_{r=1}^{R}U_1(i_1,r)\:U_2(i_2,r)\cdots U_d(i_d,r),
\end{equation}
where $A$ is a $d$-dimensional tensor, $r$ is the canonical rank and $U_j$ is the latent factor in scalar representation. Each rank-one component of the decomposition serves as a latent concept or cluster in the data. The latent factors can be interpreted as soft membership to the $r$-th latent cluster. CP is unique up to scaling and permutation of the $r$ components under very mild conditions, i.e., the components should be ``sufficiently different'' and their number not unreasonably large. Although CP format bypasses the curse of dimensionality, CP approximation may involve numerical instabilities for very high-order tensors because the problem is generally ill-posed due to intrinsic uncloseness.

\emph{Tucker decomposition (TD)} captures the interactions between the latent factors $U_i$ using a core tensor $G$ that reflects the main subspace variation in each mode assuming a multilinear structure, TD is in the form of \begin{equation}
\begin{split}
A(i_1,\cdots,i_d)\cong&\sum_{r_1=1}^{R_1}\sum_{r_2=1}^{R_2}\cdots\sum_{r_d=1}^{R_d}G(r_1,r_2,\cdots,r_d) \\
& U_1(i_1,r_1)\:U_2(i_2,r_2)\cdots U_d(i_d,r_d)
\end{split}
\label{eq:td}
\end{equation}
TD is non-unique because the latent factors can be rotated without affecting the reconstruction error. TD yields a good low-rank approximation of a tensor, since the core tensor $G$ is the best compression of the original tensor with respect to squared error. However, Tucker format is not practical for tensor order $d>5$ because the number of entries of the core tensor scales exponentially with $d$.

\emph{Hierarchical Tucker (HT) decomposition}~\cite{hackbusch2009, grasedyck2010} approximates well high-order tensors ($d>>3$) without suffering from the curse of dimensionality. HT requires a priori knowledge of a binary tree of matricizations of the tensor, 
\begin{equation}
\begin{split}
A(i_1,\cdots,i_d)\cong& \sum_{r_{u_0}=1}^{R_{u_0}}\sum_{r_{v_0}=1}^{R_{v_0}}B_{(12\cdots d)}(r_{u_0},r_{v_0})\\
&f_{u_0}(i_{u_0},r_{u_0}) f_{v_0}(i_{v_0},r_{v_0})\\
f_{t}(i_t,r_t)\cong\sum_{r_{u}=1}^{R_{u}}&\sum_{r_{v}=1}^{R_{v}}B_{t}{(r_{u},r_{v},r_{t})} f_{u}(i_{u},r_{u}) f_{v}(i_{v},r_{v})
\end{split}
\end{equation}
where $B_t$ are ``transfer" core tensors (internal nodes), $f_u$ and $f_v$ are the corresponding left and right child nodes respectively. The leaf nodes contain the latent factors.

\begin{figure}[t]
\begin{center}
\includegraphics[width=0.8\linewidth]{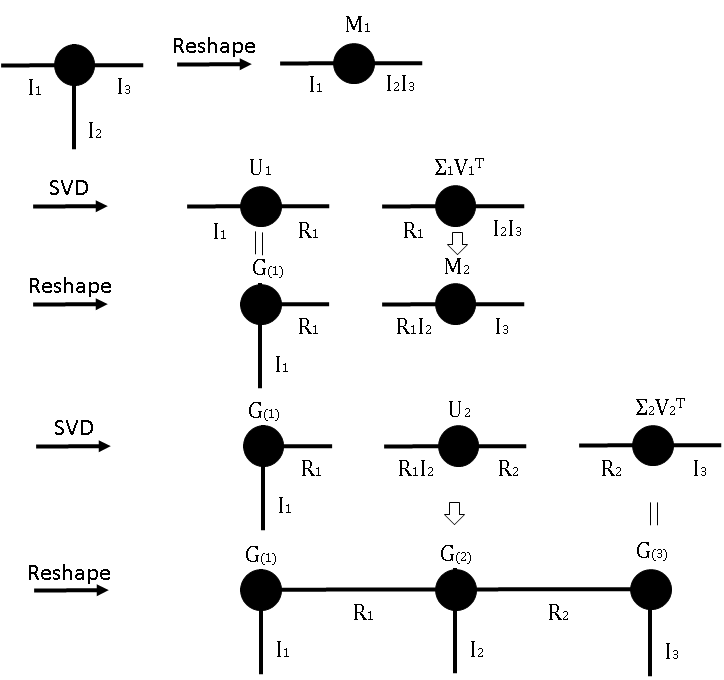}
\end{center}
   \caption{The TT-SVD algorithm for TT decomposition of a 3rd order tensor. $M_k$ is the matricization of the subtensors. The ordering of indices $I_k$ should be symmetric to get consistent SVD analyses (e.g. decay rate of singular values). For RGB color images, $I_1$: row indices, $I_2$: channels, and $I_3$: column indices. The decay rate is averaged over the sequences of SVD decomposition.}
\label{fig:tt-svd}
\end{figure}

\emph{Tensor-Train (TT) decomposition}~\cite{oseledets2011,oseledets2010tt} decomposes a given tensor into a matrix, followed by a series of three-mode ``transfer'' core tensors, and finally ended by a matrix. Each one of the core tensors is ``connected'' with its neighboring core tensors through a common reduced mode or so-called TT-rank $r_k$ with $r_0=r_d=1$. TT is given by
\begin{equation}
\begin{split}
A(i_1,\cdots,i_d)\cong&\sum_{r_1=1}^{R_1}\sum_{r_2=1}^{R_2}\cdots\sum_{r_{d-1}=1}^{R_{d-1}}G_1(r_0,i_1,r_1) \\
&G_2(r_1,i_2,r_2)\cdots G_d(r_{d-1},i_d,r_d)
\end{split}
\end{equation}
CP and TD are globally-additive models, which means tensors are represented by a (global) sum over few separable (rank-1) elements; whereas TT format is locally-multiplicative type, variables only interact directly with few local neighbors (slightly-entangled systems) through the contracted product representations. Table~\ref{tab:storage_comp} tabulates the storage complexity and storage bound of different TN. TT format exhibits both very good numerical properties and allows control of the approximation error within the decomposition algorithm. Fig.~\ref{fig:tt-svd} shows the TT-SVD algorithm for TT decomposition and Section~\ref{subsec:robust_ttsvd} explains the basics of SVD. Mathematical operations in TT format increase the TT-ranks, TT-rounding algorithm, which is mathematically similar to TT-SVD but in TT format, can efficiently reduce the TT-ranks to optimal. 
\begin{table}
\begin{center}
\begin{tabular}{ |c|c|c| } 
 \hline
 \textbf{TN}&\textbf{Storage Complexity}&\textbf{Storage Bound} \\\hline 
 CP &$\sum_{k=1}^d I_k R$&$O(dIR)$ \\\hline 
 TD &$\sum_{k=1}^d I_k R_k+\prod_{k=1}^d R_k$&$O(dIR+R^d)$ \\\hline
 HT &$\sum_{k=1}^d I_k R_k + \sum_{u,v,t} R_u R_v R_t$&$O(dIR+dR^3)$ \\\hline
 TT &$\sum_{k=1}^{d}I_k R_{k-1} R_{k}$&$O(dIR^2)$ \\\hline
\end{tabular}
\end{center}
\caption{Storage complexity of TN~\cite{Cichocki2016}. $d$ is the tensor order, $I_k$ and $R_k$ are the size and rank of mode $k$ respectively. The storage bound is calculated by letting $I=\max_k I_k$ and $R=\max_k R_k$ for all possible $k$ in particular TN.}
\label{tab:storage_comp}
\end{table}

\subsection{Robust TT-SVD Algorithm}
\label{subsec:robust_ttsvd}
\begin{figure}
\centering
\includegraphics[width=1\linewidth]{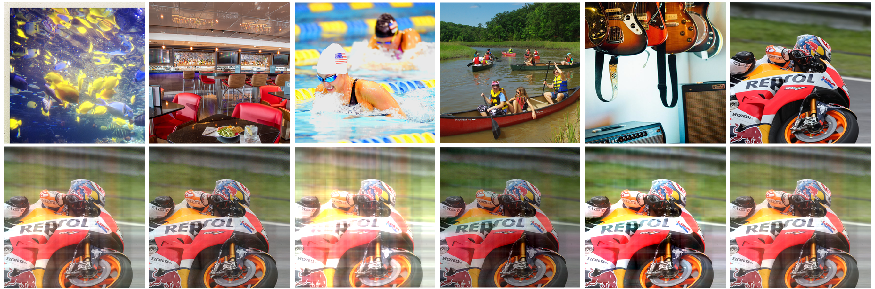}
\caption{The effect of transferring singular values between images. Last image from top row shows the original image of a motorbike. Bottom row shows the change of luminance / texture after the transfer of singular value distribution from the top images. Last image on the bottom row shows the transfer of average of all the singular values of the top images.}
\label{fig:tt_change_sing_val}
\end{figure}

\begin{figure}
\centering
\includegraphics[width=1\linewidth]{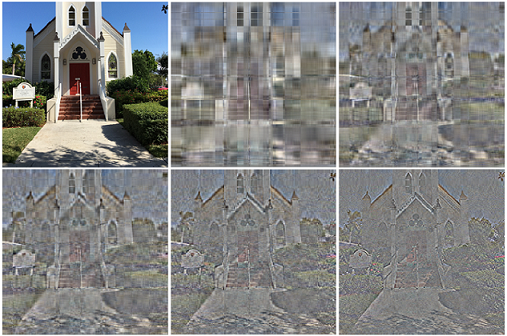}
\caption{(Left to right, top to bottom) The original and reconstructed images by combining the left and right singular vectors from 10, 20, 30, 50, and 100 largest singular values. Singular vectors encode the multiscale correlation structure of the original image. It can be observed that large singular values are associated with large-scale variation (low frequency components) and vice versa for fine-scale variation (high frequency components).}
\label{fig:tt_change_sing_vec}
\end{figure}

Singular value decomposition (SVD) decomposes a matrix $A$ into left and right singular vectors, the basis vectors are ranked by the amount of explained variation in $A$ or the so-called singular values. Mathematically, SVD is given by $A\cong U\Sigma V^T$,
where $U$ and $V$ are orthonormal matrices that contain the left and right singular vectors in their respective columns, the diagonal elements of $\Sigma$ matrix contain the corresponding singular values. As shown in Fig.~\ref{fig:tt_change_sing_val} and~\ref{fig:tt_change_sing_vec}, the distribution of singular values affect the luminance variation that accounts for textural changes such as smoothness / roughness change; the singular vectors form the basis images that encode the structural information of the original image~\cite{narwaria2012svd}. The TT-SVD algorithm in Fig.~\ref{fig:tt-svd} is used to extend SVD analyses to higher-order tensors such as color images. Unlike discrete Fourier, cosine, or wavelet transform, the basis images of SVD are not fixed but adaptively-derived, thus allows better representation of the image structure. Perturbation theory of SVD shows that the closeness of a singular value from its neighbors controls the sensitivity of its singular vector to perturbations~\cite{stewart1990stochastic}. It is further shown that when two singular values are close enough, the corresponding singular vectors are not unique, approximation of the change in subspace spanned by the corresponding singular vectors cannot be done with first-order perturbation theory but requires higher-order terms~\cite{liu2008first}. The singular values of real-life images follow exponential decay distribution, therefore the decay rate provides a scale-independent measure of the closeness of singular values and allows comparison of the robustness of multiscale correlation structure to perturbations between different datasets and image-processing techniques.

Unlike SVD, the transform kernels of convolutional neural networks (CNN) are learned from data and fixed after the training is complete; the CNN filters are not constrained to be orthogonal, but the domain of possible filter choices for both SVD and CNN spans the input space. Once training is complete, the input subspace spanned by CNN's filters is a subset of the whole input space. To make it more precise, suppose we split the input into irrelevant and relevant parts $A+\Delta{A}$ that activate particular CNN's neuron, both parts share a subset of all the singular vectors, i.e., $A+\Delta{A}=U(\Sigma+\Delta\Sigma) V^T$. In the case that $\vert\vert\Sigma\vert\vert>>\vert\vert\Delta\Sigma\vert\vert$, the robustness of the subspace spanned by the corresponding singular vectors of $\Delta\Sigma$ is determined by the closeness of the set of singular values and their neighbors. There are a few loss terms in CNN: the approximation loss due to limited number of transform kernels, the rectification loss due to nonlinear activations, spatial pooling, and dropout layers. The theoretical foundations of CNN in subspace approximation are laid down by Kuo~\cite{kuo2017cnn,kuo2018data,kuo2018interpretable,chen2018saak}. The research on subspace-based signal analysis using SVD is well-established among the signal processing community, in particular the sensitivity of subspace approximation with individual singular vectors when the singular values are close~\cite{van1993subspace,liu2008first}. If the singular values are well-separated, it can be shown that both the singular values and vectors change in the order of noise magnitude~\cite{stewart1998perturbation}. As will be shown in Section~\ref{sec:exp} Experiments, the decay rate of real-life images is small, therefore we hypothesize that the close separation between singular values gives rise to the adversarial examples in deep learning models, i.e., the neuron's activation patterns are not unique and extremely sensitive to input perturbations. 

Additionally, we propose a robust SVD algorithm (Algorithm~\ref{alg:robust_SVD}) to generate quasi-distinct singular values from closely-separated ones. In doing so, the resultant singular vectors are more robust to input perturbations, the reconstructed images are used to detect strong and static adversarial attacks. First, the cumulative sum in ascending order of the singular values, $\hat{S}$ is divided into multiple bins with exponentially-distributed bin edges. The singular values and corresponding singular vectors that fall within a bin are summed up and linearly-combined respectively. This is in accordance with the theory of degenerate matrix SVD such that any normalized linear combination of singular vectors that share the same singular value is a valid singular vector of that singular value. By merging with TT-SVD (see Fig.~\ref{fig:tt-svd}), the robust SVD algorithm can be extended to higher-order tensors such as 2D / 3D color images, videos, and hyperspectral images. The algorithm is efficient because the computational complexity increases linearly with the tensor mode size.

\begin{algorithm}
\begin{algorithmic}
\STATE \textbf{input:} matrix $A$.
\STATE \textbf{parameter:} $\alpha$, $\beta$ of exponential distribution.
\STATE \textbf{output:} quasi-distinct singular values $S$ and corresponding (linearly-combined) singular vectors $U$, $V$.
\STATE 
\STATE Initialize $\hat{S}_{binEdges}\leftarrow[0,\hspace{2pt}\alpha\exp{(\beta z)}]$, $z\in{\{0, 1, \cdots\}}$
\STATE \textbf{begin}
\STATE \hspace{10pt} $[U_0, S_0, V_0]\leftarrow svd(A)$
\STATE \hspace{10pt} $\hat{S}_0\leftarrow cumsum(S_0,`reverse')$ 
\STATE \hspace{10pt} $index\leftarrow bucketize(\hat{S}_0, \hat{S}_{binEdges})$
\STATE \hspace{10pt} $S\leftarrow accumArray(S_0, index, `sum')$
\STATE \hspace{10pt} $U\leftarrow accumArray(U_0, index, `average')$
\STATE \hspace{10pt} $V\leftarrow accumArray(V_0, index, `average')$
\STATE \hspace{10pt} Rearrange $S, U, V$ in descending order of $S$
\STATE \hspace{10pt} Normalize $U\leftarrow\frac{U}{\vert\vert U \vert\vert_2}, V\leftarrow\frac{V}{\vert\vert V \vert\vert_2}$
\STATE \hspace{10pt} (Optional) SVD rank truncation
\STATE \hspace{10pt} Return $S, U, V$
\STATE \textbf{end}
\caption{Robust SVD (can replace SVD in TT-SVD or TT-rounding algorithms, see Fig.~\ref{fig:tt-svd} for TT-SVD)}
\label{alg:robust_SVD}
\end{algorithmic}
\end{algorithm}


\section{Adversarial Attacks and Defenses}
\label{sec:adv}
 Adversarial attacks can be targeted or untargeted; the choice / structure imposed on the input perturbations is typically shaped by an $\ell_p$-norm distance metric and computed with gradient-based or optimization-based techniques with the objective to decrease the model performance. In our study, the adversarial strength is measured by normalized $\ell_2$-dissimilarity~\cite{guo2017countering}, the adversary is assumed to know only the CNN models but has no ability to influence them. 

Adversarial defenses can be broadly categorized into adversarial training and input transformations. Adversarial training requires prior knowledge of the type of attacks and train the models to differentiate them, therefore the amount of computational cost is much higher. Input transformations use either traditional image-processing techniques or generative models to remove adversarial examples; the technique is less expensive but susceptible to adaptive attacks who know the transformation techniques. We focus on input transformations that have been used against adversarial attacks in previous studies and explain the reason of their effectiveness based on our proposed theory.


\section{Experiments}
\label{sec:exp}
\emph{Datasets.} MNIST~\cite{Lecun1998} is a widely used dataset for digit classification that was introduced in 1998. It consists of $28\times 28$ pixel grayscale images of handwritten digits. There are $10$ classes ($10$ digits), $60,000$ training images, and $10,000$ testing images. Street View House Numbers (SVHN)~\cite{netzer2011reading} is an MNIST-like $32\times 32$ pixel color images consists of $73,257$ training images and $26032$ testing images for $10$ classes ($10$ digits). They are taken from Google Street View images and usually corrupted by natural phenomena like severe blur, distortion, and illumination effects on top of wide style and font variations~\cite{netzer2011reading}. CIFAR-10~\cite{Krizhevsky2009} is a dataset released in 2009 that consists of $32\times 32$ pixel color images of 10 mutually exclusive classes with $50,000$ training images and $10,000$ test images. ImageNet~\cite{russakovsky2015imagenet} is used for large-scale evaluation, it was first introduced in 2010 and the dataset stabilized in 2012. ImageNet contains color images of at least $256\times 256$ pixel with $1000$ classes; only the validation set consists of $50,000$ images ($50$ per class) are used. Section~\ref{subsec: dim_red} benchmarks the TN storage complexity, algorithmic efficiency, and model performance using these datasets. The privacy and security issues in image recognition are studied using $1000$ development (color) images of $299\times 299$ pixel released in NIPS 2017 Adversarial Attacks and Defenses Competition. This dataset is referred to as ``ImageNet"~\cite{russakovsky2015imagenet} in Section~\ref{subsec: privacy}, \ref{subsec:adv_robust}, and \ref{subsec:adv_detect_defend} due to their similar task difficulty.

\emph{Experimental Setup.} To compute the TN decomposition in Section~\ref{subsec: dim_red}, Matlab 2017a and several toolboxes are used: Tensorlab 3.0~\cite{vervliet2016tensorlab}, htucker 1.2~\cite{kressner2012htucker}, and TT-toolbox 2.2.2. The compression and decompression time are benchmarked using Intel(R) Xeon(R) CPU E5-1650 v4 @ 3.60GHz 3.60GHz. CP and TD are computed using Tensorlab ``cpd\_gevd" and ``mlsvd" functions, HT using htucker ``htensor.truncate\_ltr" function, and TT using TT-toolbox ``tt\_tensor" function. These functions use generalized eigenvalue or SVD to speed up the decomposition. Adversarial defenses based on input transformation using color bit-depth reduction~\cite{xu2017feature}, cropping-rescaling, median, gaussian, and non-local means~\cite{buades2005non} filters are coded using Matlab functions and toolbox. Image quilting~\cite{efros2001image}, total variance minimization (TVM)~\cite{rudin1992nonlinear}, and JPEG compression~\cite{dziugaite2016study} are Python implementations by Guo~\etal~\cite{guo2017countering}. Adversarial attacks are generated using Tensorflow 1.4.0~\cite{abadi2016tensorflow,tensorflow2015-whitepaper} and Cleverhans v2.1.0~\cite{Papernot2016}. The CNN model for MNIST, SVHN, and CIFAR-10 is an all convolutional net~\cite{springenberg2014striving} taken from Cleverhans model zoo. The ImageNet classification is using Inception v3 ~\cite{szegedy2016rethinking}. Fast Gradient Method (FGM)~\cite{goodfellow2015explaining}, Basic Iteration Method (BIM)~\cite{kurakin2016adversarial}, and Deep Fool (DF)~\cite{moosavi2016deepfool} are gradient-based attacks whereas Carlini-Wagner (CW)~\cite{carlini2017towards} and Elastic Net Method (EAD)~\cite{chen2017ead} are optimization-based adversarial attacks. We follow closely the method proposed by Guo~\etal~\cite{guo2017countering} to generate adversarial examples with increasing adversarial strength. FGM and BIM are done by adjusting the hyperparameters; DF and CW perturbations are amplified to increase the normalized $\ell_2$-dissimilarity after successful attacks. Universal Perturbations (UP)~\cite{moosavi2017universal} and Adversarial Patch (AP)~\cite{brown2017adversarial} are strong static attacks which can be image- and network-agnostic. Only the image-agnostic case is considered here. UP is generated using BIM in each iteration; whereas AP is taken from the implementations by Brown~\etal~\cite{brown2017adversarial}. Different from other adversarial techniques that manipulate pixels within $\ell_p$ distance which may sometimes produce noticeable artifacts, spatially transformed adversarial example (stAdv)~\cite{xiao2018spatially} is a new approach that generates realistic adversarial examples with smooth image deformation; their code is made publicly available by Dumont~\etal\cite{dumont2018robustness}. In our study, AP and stAdv are targeted attacks with ``toaster" and randomized targets respectively, others are untargeted attacks. Because stAdv deforms images smoothly, which stands in contrast to our proposed theory that adversarial examples tend to increase the image roughness, we report the stAdv hyperparameter that regularizes the local distortion characterized by a flow field and adversarial loss used in our study, i.e., MNIST ($10^{-2}$), SVHN ($10^{-3}$), CIFAR-10 ($10^{-3}$), and ImageNet ($10^{-6}$). Notice that the regularization decreases with dataset complexity, this means that it is much harder to generate adversarial examples with smooth deformation for complex images.

\subsection{Dimensionality Reduction of Input Data}
\label{subsec: dim_red}
\begin{table}
\begin{center}
\begin{tabular}{ |c|c|c|c| } 
 \hline
 \textbf{TN}&\textbf{Time (ms)}&\textbf{Compression}&\textbf{Top-1}\\
  &&\textbf{Ratio}&\textbf{Accuracy}\\\hline
  \multicolumn{3}{|c|}{\textbf{MNIST Dataset}}&0.993\\\hline
  CP&N/A&N/A&N/A\\\hline
  TD&0.6 / 0.3&0.390&0.988\\\hline
  HT&2.5 / 0.5&0.422&0.988\\\hline
  TT&0.39 / 0.07&0.439&0.990\\\hline
  \multicolumn{3}{|c|}{\textbf{SVHN Dataset}}&0.951\\\hline
  CP&7.9 / 0.14&0.327&0.945\\\hline
  TD&1.6 / 0.37&0.309&0.950\\\hline
  HT&2.8 / 0.56&0.342&0.950\\\hline
  TT&0.97 / 0.07&0.150&0.929\\\hline
  \multicolumn{3}{|c|}{\textbf{CIFAR-10 Dataset}}&0.858\\\hline
  CP&7.8 / 0.13&0.334&0.811\\\hline
  TD&1.6 / 0.39&0.309&0.8105\\\hline
  HT&2.9 / 0.58&0.342&0.809\\\hline
  TT&1.2 / 0.09&0.455&0.833\\\hline
  \multicolumn{3}{|c|}{\textbf{ImageNet Dataset}}&0.748\\\hline
  CP&214 / 2.2&0.460&0.535\\\hline
  TD&53.9 / 4.3&0.335&0.693\\\hline
  HT&53.8 / 3.2&0.372&0.703\\\hline
  TT&48.3 / 2.7&0.501&0.655\\\hline
\end{tabular}
\end{center}
\caption{Model performance using TN for input data compression. The time for compression / decompression per image is measured in milliseconds. CP decomposition is not available (N/A) for MNIST dataset because of algorithmic instability. This happens occasionally for other datasets, hence the original data and size are used instead for calculation of compression ratio.}
\label{tab:dim_reduce}
\end{table}
Table~\ref{tab:dim_reduce} benchmarks the compression / decompression time for TN with different compression ratio. The time needed generally increases with TN size. CP decomposition is about $4\times$ longer than other TN decomposition. TN decompression time is generally much faster compared to compression time. After TN decomposition, the cores and latent factors are quantized to 8-bit depth. Some of the subtensors can be uniformly quantized but some requires non-uniform quantization using Lloyd's algorithm~\cite{lloyd1982least,max1960quantizing} to reduce the image distortion, e.g., TD's core $G$ in Eq.~\ref{eq:td}. It can be observed that TN generally retains the features for image classification by CNNs without the need to retrain the model; at least half of the storage size can be saved using TN for data compression.

\subsection{Privacy-Preserving Distributed Data Storage and Communication}
\begin{figure}
\centering
\includegraphics[width=0.8\linewidth]{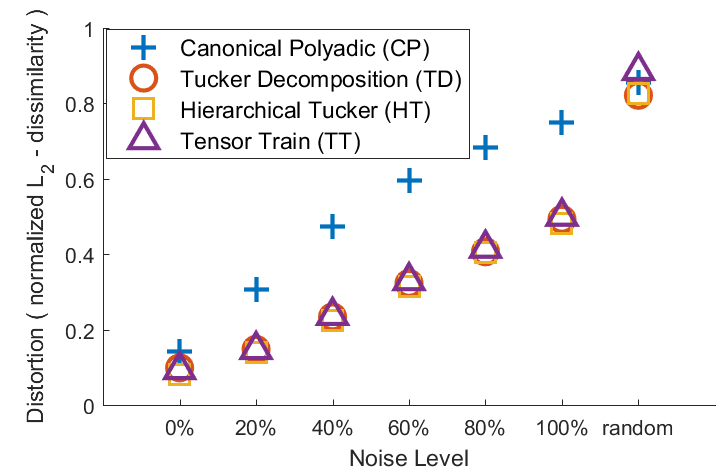}
\caption{Image distortion resulted from adding noise to a randomly-selected core of the TN. Note that ``random" label in the x-axis means randomize the sequence in the selected core.}
\label{fig:tn_noise_distort}
\end{figure}
\label{subsec: privacy}
\begin{figure}
\centering
\includegraphics[width=1\linewidth]{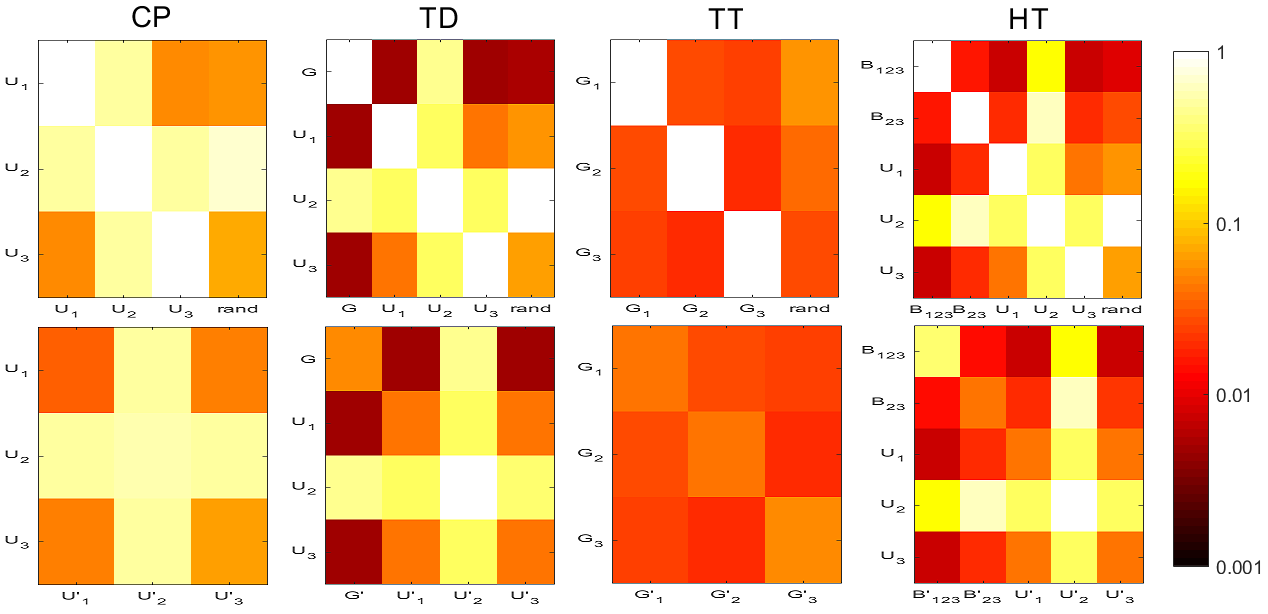}
\caption{Normalized mutual information between cores and latent factors of one image (top row) and two different images (bottom row) for different TNs. Note that ``rand" label in the x-axis means cores with uniformly-distributed noise.}
\label{fig:tn_corr_map}
\end{figure}

Fig.~\ref{fig:tn_noise_distort} shows the image distortion as a result of adding noise to randomly-selected TN subtensor, the effect is larger if the perturbations is applied on the singular vectors corresponding to the leading singular values, however this information is usually unknown to the adversary. CP's distortion is larger because the format is more compact compared to other TNs. Due to the diverse possible topology structure of decomposition for a given tensor, Wang~\etal\cite{wang2018big} proposes three different security models to process data generated by cyber-physical-social systems, i.e., open model, half-open model, and encrpted model to process data with different level of sensitivity and privacy requirements. The tensor formats and topology structure are made private to selected users for half-open and encrypted models. Here, we experimentally verify that the neighboring subtensors could not be identified based on information theory. Mutual information is commonly used to cross-examine the information content between subtensors~\cite{Cichocki2016}. Fig.~\ref{fig:tn_corr_map} shows the normalized mutual information (NMI) between two subtensors of particular TN for one image, two images, and random noise, the results show that they are indistinguishable from each other. NMI is a universal metric such that any other distance measure judges two random variables close-by, NMI will also judge them close. As shown in Fig.~\ref{fig:tn_corr_map}. the NMI variation is largely attributed to the variation in subtensor's value distribution, if the variation in particular subtensor is high (i.e., entropy is high), its NMI with other subtensor is likely to be smaller.  

\subsection{Robustness against Adversarial Attacks}
\begin{figure}
\centering
\includegraphics[width=1.0\linewidth]{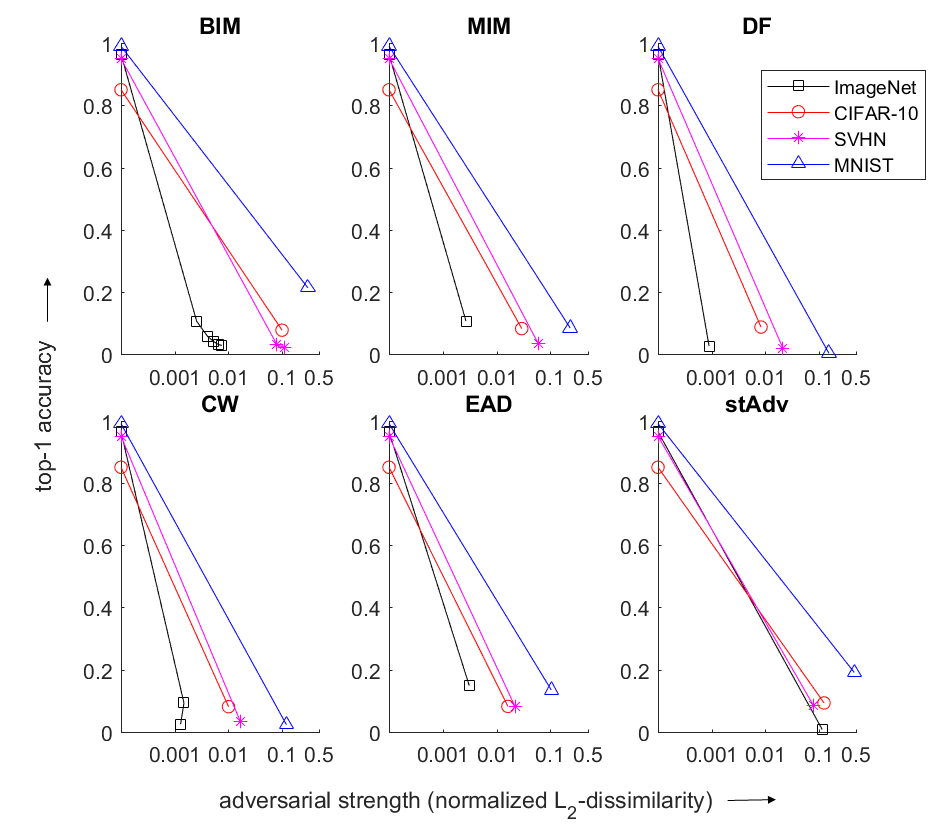}
   \caption{Model accuracy of datasets under adversarial attacks with increasing adversarial strength measured in normalized $\ell_2$-dissimilarity. Notice the robustness of the datasets to adversarial attacks, i.e., MNIST $>$ SVHN $\gtrsim$ CIFAR-10 $>$ ImageNet.}
\label{fig:top1_l2diss}
\end{figure}

The decay rate of leading singular values determines the robustness of subspace approximation by CNNs. Table~\ref{tab:dataset_slope} tabulates the decay rate for 1000 randomly-selected images from MNIST, SVHN, CIFAR-10, and ImageNet datasets using the 5th-25th largest singular values. Coincidentally, the decay rate correlates well to the datasets' complexity. Fig.~\ref{fig:top1_l2diss} shows the robustness of the datasets to adversarial attacks. It can be observed that the steeper the dataset's TT-SVD slope, the more robust the dataset to a wide variety of different adversarial attacks. In particular, input perturbations by stAdv is done by smooth deformation; it requires much higher adversarial strength for successful attacks compared to other techniques. This agrees with our theory that adversarial examples tend to decrease the decay rate of singular values to increase sensitivity of the subspace approximation by CNNs, hence increase the image roughness as a result. Table~\ref{tab:attack_slope} shows that strong adversarial attacks flatten the TT-SVD slope. Effectiveness of defenses based on input transformation has been studied before, our theory explains the reason why spatial smoothing techniques provide more resistance to adversarial attacks, as shown in~\cite{guo2017countering,xu2017feature}. This is because spatial smoothing steepen the TT-SVD slope (see Table~\ref{tab:defend_slope}), hence reduce the sensitivity of subspace approximation by CNNs.
\label{subsec:adv_robust}
\begin{table}
\begin{center}
\begin{tabular}{ |c|c|c| } 
 \hline
 \textbf{Datasets}&\textbf{Image Size}&\textbf{TT-SVD Slope}\\\hline
 MNIST (grayscale)&28x28x1&$-0.4\pm0.5$\\\hline
SVHN (color)&32x32x3&$-0.38\pm0.12$\\\hline
CIFAR10 (color)&32x32x3&$-0.17\pm0.04$\\\hline
ImageNet (color)&299x299x3&$-0.072\pm0.019$\\\hline
ImageNet (color)&299x299x3&$-0.061\pm0.016$\\
+ 30\% noise&&\\\hline
\end{tabular}
\end{center}
\caption{Robustness of correlation structure of different datasets measured by the TT-SVD slope. Steeper slope means the separation between singular values are larger, hence the subspace approximation by CNNs is more robust to input perturbations. The standard deviation of the TT-SVD slope measures the variability of the estimated value. Notice that adding noise flattens the TT-SVD slope, hence decreases the robustness of correlation structure.}
\label{tab:dataset_slope}
\end{table}
\begin{table}
\begin{center}
\begin{tabular}{ |c|c|c| } 
 \hline
 \textbf{Adversarial}&\textbf{Top-1}&\textbf{TT-SVD}\\
 \textbf{Attacks (Strength)}&\textbf{Accuracy}&\textbf{Slope}\\\hline
FGM, $\ell_\infty$ (0.0812)&0.25&$-0.068\pm0.017\boldsymbol\uparrow$\\\hline
FGM, $\ell_2$ (0.0773)&0.25&$-0.067\pm0.016\boldsymbol\uparrow$\\\hline
BIM, $\ell_\infty$ (0.0778)&0.016&$-0.069\pm0.017\boldsymbol\uparrow$\\\hline
BIM, $\ell_2$ (0.0784)&0.014&$-0.068\pm0.016\boldsymbol\uparrow$\\\hline
CW, $\ell_2$ (0.0775)&0.17&$-0.067\pm0.017\boldsymbol\uparrow$\\\hline
DF, $\ell_2$ (0.0765)&0.134&$-0.066\pm0.016\boldsymbol\uparrow$\\\hline
UP, $\ell_\infty$ (0.0832)&0.196&$-0.068\pm0.017\boldsymbol\uparrow$\\\hline
UP, $\ell_2$ (0.0787)&0.196&$-0.068\pm0.017\boldsymbol\uparrow$\\\hline
AP (0.64)&0.0&$-0.062\pm0.008\boldsymbol\uparrow$\\\hline
\end{tabular}
\end{center}
\caption{Similar to Table~\ref{tab:dataset_slope} but for adversarial attacks. Adversarial strength is measured by normalized $\ell_2$-dissimilarity. The upwards arrow means that the technique flattens the TT-SVD slope by more than $10\%$ of the slope variability, vice versa for downwards arrow. The original slope value is $-0.072\pm0.019$.}
\label{tab:attack_slope}
\end{table}
\begin{table}
\begin{center}
\begin{tabular}{ |c|c|c| } 
 \hline
 \multicolumn{2}{|c|}{\textbf{Adversarial Defenses based}}&\textbf{TT-SVD Slope}\\
 \multicolumn{2}{|c|}{\textbf{on Input Transformation}}&\\\hline
\multicolumn{3}{|c|}{\textbf{Spatial Smoothing}}\\\hline
\multicolumn{2}{|l|}{Cropping-Rescaling~\cite{graese2016assessing}}&$-0.089\pm0.026\boldsymbol\downarrow$\\\hline
 \multicolumn{2}{|l|}{Image Quilting~\cite{efros2001image}}&$-0.079\pm0.019\boldsymbol\downarrow$\\\hline
 \multicolumn{2}{|l|}{Median Filter~\cite{xu2017feature}}&$-0.080\pm0.020\boldsymbol\downarrow$\\\hline
 \multicolumn{2}{|l|}{Gaussian Filter~\cite{xu2017feature}}&$-0.088\pm0.022\boldsymbol\downarrow$\\\hline
 \multicolumn{2}{|l|}{Non-Local Means Filter~\cite{buades2005non}}&$-0.081\pm0.021\boldsymbol\downarrow$\\\hline
 \multicolumn{2}{|l|}{Total Variance Minimization~\cite{rudin1992nonlinear}}&$-0.081\pm0.020\boldsymbol\downarrow$\\\hline
 \multicolumn{3}{|c|}{\textbf{Amplitude Quantization}}\\\hline
Color Bit-Depth&4-bit&$-0.071\pm0.018$\\
Reduction \cite{xu2017feature}&3-bit&$-0.069\pm0.017\boldsymbol\uparrow$\\\hline
\multicolumn{3}{|c|}{\textbf{Frequency-based Compression}}\\\hline
&\underline{quality level}&\\
JPEG~\cite{dziugaite2016study}&75&$-0.072\pm0.019$\\
Compression&50&$-0.073\pm0.019$\\
&25&$-0.073\pm0.019$\\
&5&$-0.071\pm0.017$\\\hline
\end{tabular}
\end{center}
\caption{Similar to Table \ref{tab:attack_slope} but for adversarial defenses.}
\label{tab:defend_slope}
\end{table}

\subsection{Detect Strong and Static Adversarial Attacks}
\label{subsec:adv_detect_defend}
As shown in Table~\ref{tab:attack_detection}, the detection of strong static attacks using our proposed robust TT-SVD algorithm only requires bounding the $\ell_2$-norm between image before and after reconstruction. Currently, the proposed algorithm works well if the image consists of ``simple" correlation structure (high SVD's decay rate), e.g., single object recognition. Images with complex variation or cluttered scene may consider pre-processing with spatial smoothing and cropping-rescaling ~\cite{graese2016assessing} respectively before using the proposed algorithm. Existing adversarial detection techniques rely on (1) sample statistics, (2) prediction inconsistency, and (3) training a detector; (1) is ineffective, (2) needs to process a batch of images each time, (3) needs to have labelled data and model training takes time~\cite{xu2017feature}. The robust TT-SVD algorithm provides a new way to detect adversarial examples directly on the input; the algorithm is adaptive in nature because the singular values / vectors are adaptively-derived. 

\begin{table}
\begin{center}
\begin{tabular}{ |c|c|c|c| } 
 \hline
 \textbf{Adversarial}&\textbf{Norm}&\textbf{Adversarial}&\textbf{Detection}\\
 \textbf{Attacks}&&\textbf{Strength}&\textbf{Rate}\\\hline
FGM~\cite{goodfellow2015explaining} &$\ell_\infty$&0.071&0.997\\
&$\ell_2$&0.043&0.956\\\hline
BIM~\cite{kurakin2016adversarial}&$\ell_\infty$&0.069&0.997\\
&$\ell_2$&0.048&0.985\\\hline
CW~\cite{carlini2017towards} &$\ell_2$&0.046&0.944\\\hline
DF~\cite{moosavi2016deepfool} &$\ell_2$&0.037&0.932\\\hline
UP~\cite{moosavi2017universal} &$\ell_\infty$&0.052&0.997\\
&$\ell_2$&0.056&0.994\\\hline
AP~\cite{brown2017adversarial}&-&$0.045$&0.991\\\hline
\end{tabular}
\end{center}
\caption{Detection rate of strong and static adversarial attacks. The adversarial strength is measured by normalized $\ell_2$-dissimilarity.  The slope of cumulative distribution of TT-SVD is set to $-0.03$ and truncation error $\leqslant0.03$. $337$ out of $1000$ development images from NIPS 2017 Adversarial Attacks and Defenses Competition are selected by setting the initial $\ell_2$-norm $<1000$. The $\ell_2$-dissimilarity of the $337$ samples is $0.01$. Adversarial attacks are detected when $\ell_2$-norm $>1000$.}
\label{tab:attack_detection}
\end{table}

\section{Related Work}

Tensor decomposition has been used for feature extraction and classification on multidimensional data~\cite{phan2010}, recent work is extended to high-dimensional data with cutting-edge tensor techniques such as tensor-train decomposition~\cite{bengua2017,bengua2015}. Theoretical links between tensor networks and deep learning are slowly establishing, for example, the expressive power of CP and HT decompositions correspond to shallow and deep networks respectively~\cite{cohen2016,cohen2016a}, whereas TT corresponds to recurrent neural network~\cite{khrulkov2017}. Tensor networks have been used to compress and accelerate deep learning models due to the high redundancy in model parameters, e.g., fully-connected network~\cite{Novikov2015}, convolutional network~\cite{kim2015,tai2015,lebedev2014}, recurrent network~\cite{tjandra2017compressing}, sharing residual units~\cite{yunpeng2017}, multitask learning~\cite{yang2016}, multimodal data~\cite{zhang2018tensor,zhang2017tucker,zhang2017improved,chien2017tensor,kossaifi2017tensor}. Additionally, tensor power technique is used to learn latent variable models efficiently with statistically consistent estimator~\cite{anandkumar2014,janzamin2015}. Due to the versatility of tensor representations, tensor techniques have been proposed for big data networking and management~\cite{yu2017,kuang2016a,Kuang2016b,yang2015,kuang2014}, e.g., Internet of Things~\cite{acar2016,kuang2016}. Privacy-preserving techniques for tensor decomposition have also been studied~\cite{wang2016,kuang2015} for cloud computing. Exploring tensor networks as an alternative data structure for efficient and distributed storage as well as privacy preservation for input data of deep learning or other machine learning models have not be been considered before; our study is particularly relevant in the context of edge and fog computing. We exploit the data redundancy and uses tensor network decomposition to retain the relevant features for image classification.

In the field of computer vision, SVD has been used for image denoising, compression, and forensic such as steganography and watermarking. Higher-order SVD can help to disentangle the constituents factors or modes of image ensembles, e.g., TensorFaces~\cite{vasilescu2002multilinear} for facial images with different lighting conditions, viewpoint and poses. Closely related to our work is the use of SVD to extract features for visual quality assessment. Reference images are usually provided for comparison between images before and after processing~\cite{narwaria2012svd}, whereas no reference measure requires assumptions on the patches to be analyzed such as anisotropy~\cite{zhu2010automatic}; all of the SVD-based image quality metrics work on grayscale images. Adversary does not provide the reference images for comparison; our work extends the SVD properties to higher-order tensors and analyze the robustness of correlation structure extracted from input data for image classification by CNNs. 

\section{Conclusion}

At first glance, it may seems obvious that adversarial attacks increase the image roughness, therefore natural choices for adversarial defense based on input transformation should incorporate different levels of spatial smoothing, e.g., local, non-local, edge-preserving, etc. Further experiments show that the robustness against adversarial attacks differ between datasets with different decay rate of SVD singular values. This suggests that there is a deeper level connections between adversarial examples in deep learning models and SVD's decay rate. Perturbation theory of SVD shows that the closeness between singular values controls the sensitivity of subspace approximation by CNNs. Empirical results show that real-life images typically have slow SVD's decay rate, which explains the cause of adversarial examples in CNNs. The subspace approximation is more stable to input perturbations if the CNN's loss from approximation, pooling, dropout, and rectification can be reduced.

{\small
\bibliographystyle{ieee}
\bibliography{tn_dl}

\begin{thebibliography}{10}\itemsep=-1pt

\bibitem{tensorflow2015-whitepaper}
M.~Abadi, A.~Agarwal, P.~Barham, E.~Brevdo, Z.~Chen, C.~Citro, G.~S. Corrado,
  A.~Davis, J.~Dean, M.~Devin, S.~Ghemawat, I.~Goodfellow, A.~Harp, G.~Irving,
  M.~Isard, Y.~Jia, R.~Jozefowicz, L.~Kaiser, M.~Kudlur, J.~Levenberg,
  D.~Man\'{e}, R.~Monga, S.~Moore, D.~Murray, C.~Olah, M.~Schuster, J.~Shlens,
  B.~Steiner, I.~Sutskever, K.~Talwar, P.~Tucker, V.~Vanhoucke, V.~Vasudevan,
  F.~Vi\'{e}gas, O.~Vinyals, P.~Warden, M.~Wattenberg, M.~Wicke, Y.~Yu, and
  X.~Zheng.
\newblock {TensorFlow}: Large-scale machine learning on heterogeneous systems,
  2015.
\newblock Software available from tensorflow.org.

\bibitem{abadi2016tensorflow}
M.~Abadi, P.~Barham, J.~Chen, Z.~Chen, A.~Davis, J.~Dean, M.~Devin,
  S.~Ghemawat, G.~Irving, M.~Isard, et~al.
\newblock Tensorflow: a system for large-scale machine learning.
\newblock In {\em OSDI}, volume~16, pages 265--283, 2016.

\bibitem{acar2016}
E.~Acar, A.~Anandkumar, L.~Mullin, S.~Rusitschka, and V.~Tresp.
\newblock Tensor computing for internet of things.
\newblock {\em Dagstuhl Reports}, 6:57--79, 2016.

\bibitem{akhtar2018threat}
N.~Akhtar and A.~Mian.
\newblock Threat of adversarial attacks on deep learning in computer vision: A
  survey.
\newblock {\em arXiv preprint arXiv:1801.00553}, 2018.

\bibitem{anandkumar2014}
A.~Anandkumar, R.~Ge, D.~J. Hsu, S.~M. Kakade, and M.~Telgarsky.
\newblock Tensor decompositions for learning latent variable models.
\newblock {\em Journal of Machine Learning Research}, 15(1):2773--2832, 2014.

\bibitem{bengua2017}
J.~A. Bengua, P.~N. Ho, H.~D. Tuan, and M.~N. Do.
\newblock Matrix product state for higher-order tensor compression and
  classification.
\newblock {\em IEEE Transactions on Signal Processing}, 65(15):4019--4030,
  2017.

\bibitem{bengua2015}
J.~A. Bengua, H.~N. Phien, and H.~D. Tuan.
\newblock Optimal feature extraction and classification of tensors via matrix
  product state decomposition.
\newblock In {\em Big Data (BigData Congress), 2015 IEEE International Congress
  on}, pages 669--672. IEEE, 2015.

\bibitem{brown2017adversarial}
T.~B. Brown, D.~Man{\'e}, A.~Roy, M.~Abadi, and J.~Gilmer.
\newblock Adversarial patch.
\newblock {\em In Machine Learning and Computer Security Workshop, Neural
  Information Processing Systems}, 2017.

\bibitem{buades2005non}
A.~Buades, B.~Coll, and J.-M. Morel.
\newblock A non-local algorithm for image denoising.
\newblock In {\em Computer Vision and Pattern Recognition, 2005. CVPR 2005.
  IEEE Computer Society Conference on}, volume~2, pages 60--65. IEEE, 2005.

\bibitem{carlini2017towards}
N.~Carlini and D.~Wagner.
\newblock Towards evaluating the robustness of neural networks.
\newblock In {\em 2017 IEEE Symposium on Security and Privacy (SP)}, pages
  39--57. IEEE, 2017.

\bibitem{chen2017ead}
P.-Y. Chen, Y.~Sharma, H.~Zhang, J.~Yi, and C.-J. Hsieh.
\newblock Ead: elastic-net attacks to deep neural networks via adversarial
  examples.
\newblock {\em arXiv preprint arXiv:1709.04114}, 2017.

\bibitem{chen2018saak}
Y.~Chen, Z.~Xu, S.~Cai, Y.~Lang, and C.-C.~J. Kuo.
\newblock A saak transform approach to efficient, scalable and robust
  handwritten digits recognition.
\newblock In {\em 2018 Picture Coding Symposium (PCS)}, pages 174--178. IEEE,
  2018.

\bibitem{chien2017tensor}
J.-T. Chien and Y.-T. Bao.
\newblock Tensor-factorized neural networks.
\newblock {\em IEEE transactions on neural networks and learning systems},
  2017.

\bibitem{Cichocki2016}
A.~Cichocki, N.~Lee, I.~Oseledets, A.-H. Phan, Q.~Zhao, and D.~P. Mandic.
\newblock Tensor networks for dimensionality reduction and large-scale
  optimization: Part 1 low-rank tensor decompositions.
\newblock {\em Foundations and Trends\textsuperscript{\textregistered} in
  Machine Learning}, 9(4-5):249--429, 2016.

\bibitem{Cichocki2015}
A.~Cichocki, D.~Mandic, L.~{De Lathauwer}, G.~Zhou, Q.~Zhao, C.~Caiafa, and
  H.~A. Phan.
\newblock {Tensor decompositions for signal processing applications: From
  two-way to multiway component analysis}.
\newblock {\em IEEE Signal Processing Magazine}, 32(2):145--163, 2015.

\bibitem{Cichocki2017}
A.~Cichocki, A.-H. Phan, Q.~Zhao, N.~Lee, I.~Oseledets, M.~Sugiyama, and D.~P.
  Mandic.
\newblock Tensor networks for dimensionality reduction and large-scale
  optimization: Part 2 applications and future perspectives.
\newblock {\em Foundations and Trends\textsuperscript{\textregistered} in
  Machine Learning}, 9(6):431--673, 2017.

\bibitem{cohen2016}
N.~Cohen, O.~Sharir, and A.~Shashua.
\newblock On the expressive power of deep learning: A tensor analysis.
\newblock In {\em Conference on Learning Theory}, pages 698--728, 2016.

\bibitem{cohen2016a}
N.~Cohen and A.~Shashua.
\newblock Convolutional rectifier networks as generalized tensor
  decompositions.
\newblock In {\em International Conference on Machine Learning}, pages
  955--963, 2016.

\bibitem{dumont2018robustness}
B.~Dumont, S.~Maggio, and P.~Montalvo.
\newblock Robustness of rotation-equivariant networks to adversarial
  perturbations.
\newblock {\em arXiv preprint arXiv:1802.06627}, 2018.

\bibitem{dziugaite2016study}
G.~K. Dziugaite, Z.~Ghahramani, and D.~M. Roy.
\newblock A study of the effect of jpg compression on adversarial images.
\newblock {\em arXiv preprint arXiv:1608.00853}, 2016.

\bibitem{efros2001image}
A.~A. Efros and W.~T. Freeman.
\newblock Image quilting for texture synthesis and transfer.
\newblock In {\em Proceedings of the 28th annual conference on Computer
  graphics and interactive techniques}, pages 341--346. ACM, 2001.

\bibitem{goodfellow2015explaining}
I.~Goodfellow, J.~Shlens, and C.~Szegedy.
\newblock Explaining and harnessing adversarial examples.
\newblock {\em In Proc. ICLR}, 2015.

\bibitem{graese2016assessing}
A.~Graese, A.~Rozsa, and T.~E. Boult.
\newblock Assessing threat of adversarial examples on deep neural networks.
\newblock In {\em Machine Learning and Applications (ICMLA), 2016 15th IEEE
  International Conference on}, pages 69--74. IEEE, 2016.

\bibitem{grasedyck2010}
L.~Grasedyck.
\newblock Hierarchical singular value decomposition of tensors.
\newblock {\em SIAM Journal on Matrix Analysis and Applications},
  31(4):2029--2054, 2010.

\bibitem{Grasedyck2013a}
L.~Grasedyck, D.~Kressner, and C.~Tobler.
\newblock {A literature survey of low-rank tensor approximation techniques}.
\newblock {\em GAMM Mitteilungen}, 36(1):53--78, 2013.

\bibitem{guo2017countering}
C.~Guo, M.~Rana, M.~Cisse, and L.~van~der Maaten.
\newblock Countering adversarial images using input transformations.
\newblock {\em ICLR}, 2017.

\bibitem{hackbusch2009}
W.~Hackbusch and S.~K{\"u}hn.
\newblock A new scheme for the tensor representation.
\newblock {\em Journal of Fourier analysis and applications}, 15(5):706--722,
  2009.

\bibitem{janzamin2015}
M.~Janzamin, H.~Sedghi, and A.~Anandkumar.
\newblock Beating the perils of non-convexity: Guaranteed training of neural
  networks using tensor methods.
\newblock {\em arXiv preprint arXiv:1506.08473}, 2015.

\bibitem{khrulkov2017}
V.~Khrulkov, A.~Novikov, and I.~Oseledets.
\newblock Expressive power of recurrent neural networks.
\newblock {\em arXiv preprint arXiv:1711.00811}, 2017.

\bibitem{kim2015}
Y.-D. Kim, E.~Park, S.~Yoo, T.~Choi, L.~Yang, and D.~Shin.
\newblock Compression of deep convolutional neural networks for fast and low
  power mobile applications.
\newblock {\em ICLR}, 2015.

\bibitem{kossaifi2017tensor}
J.~Kossaifi, A.~Khanna, Z.~Lipton, T.~Furlanello, and A.~Anandkumar.
\newblock Tensor contraction layers for parsimonious deep nets.
\newblock In {\em Computer Vision and Pattern Recognition Workshops (CVPRW),
  2017 IEEE Conference on}, pages 1940--1946. IEEE, 2017.

\bibitem{kressner2012htucker}
D.~Kressner and C.~Tobler.
\newblock htucker - a matlab toolbox for tensors in hierarchical tucker format.
\newblock {\em Mathicse, EPF Lausanne}, 2012.

\bibitem{Krizhevsky2009}
A.~Krizhevsky, V.~Nair, and G.~Hinton.
\newblock Learning multiple layers of features from tiny images.
\newblock 2009.

\bibitem{kuang2014}
L.~Kuang, F.~Hao, L.~T. Yang, M.~Lin, C.~Luo, and G.~Min.
\newblock A tensor-based approach for big data representation and
  dimensionality reduction.
\newblock {\em IEEE transactions on emerging topics in computing},
  2(3):280--291, 2014.

\bibitem{kuang2015}
L.~Kuang, L.~Yang, J.~Feng, and M.~Dong.
\newblock Secure tensor decomposition using fully homomorphic encryption
  scheme.
\newblock {\em IEEE Transactions on Cloud Computing}, 2015.

\bibitem{kuang2016}
L.~Kuang, L.~T. Yang, and K.~Qiu.
\newblock Tensor-based software-defined internet of things.
\newblock {\em IEEE Wireless Communications}, 23(5):84--89, 2016.

\bibitem{kuang2016a}
L.~Kuang, L.~T. Yang, S.~C. Rho, Z.~Yan, and K.~Qiu.
\newblock A tensor-based framework for software-defined cloud data center.
\newblock {\em ACM Transactions on Multimedia Computing, Communications, and
  Applications (TOMM)}, 12(5s):74, 2016.

\bibitem{Kuang2016b}
L.~Kuang, L.~T. Yang, X.~Wang, P.~Wang, and Y.~Zhao.
\newblock A tensor-based big data model for qos improvement in software defined
  networks.
\newblock {\em IEEE Network}, 30(1):30--35, 2016.

\bibitem{kuo2017cnn}
C.-C.~J. Kuo.
\newblock The cnn as a guided multilayer recos transform [lecture notes].
\newblock {\em IEEE Signal Processing Magazine}, 34(3):81--89, 2017.

\bibitem{kuo2018data}
C.-C.~J. Kuo and Y.~Chen.
\newblock On data-driven saak transform.
\newblock {\em Journal of Visual Communication and Image Representation},
  50:237--246, 2018.

\bibitem{kuo2018interpretable}
C.-C.~J. Kuo, M.~Zhang, S.~Li, J.~Duan, and Y.~Chen.
\newblock Interpretable convolutional neural networks via feedforward design.
\newblock {\em arXiv preprint arXiv:1810.02786}, 2018.

\bibitem{kurakin2016adversarial}
A.~Kurakin, I.~Goodfellow, and S.~Bengio.
\newblock Adversarial examples in the physical world.
\newblock {\em arXiv preprint arXiv:1607.02533}, 2016.

\bibitem{lebedev2014}
V.~Lebedev, Y.~Ganin, M.~Rakhuba, I.~Oseledets, and V.~Lempitsky.
\newblock Speeding-up convolutional neural networks using fine-tuned
  cp-decomposition.
\newblock {\em ICLR}, 2015.

\bibitem{lecun2015}
Y.~LeCun, Y.~Bengio, and G.~Hinton.
\newblock Deep learning.
\newblock {\em Nature}, 521(7553):436--444, 2015.

\bibitem{Lecun1998}
Y.~LeCun, C.~Cortes, and C.~J. Burges.
\newblock The mnist database of handwritten digits.
\newblock 1998.

\bibitem{liu2008first}
J.~Liu, X.~Liu, and X.~Ma.
\newblock First-order perturbation analysis of singular vectors in singular
  value decomposition.
\newblock {\em IEEE Transactions on Signal Processing}, 56(7):3044--3049, 2008.

\bibitem{lloyd1982least}
S.~Lloyd.
\newblock Least squares quantization in pcm.
\newblock {\em IEEE transactions on information theory}, 28(2):129--137, 1982.

\bibitem{max1960quantizing}
J.~Max.
\newblock Quantizing for minimum distortion.
\newblock {\em IRE Transactions on Information Theory}, 6(1):7--12, 1960.

\bibitem{moosavi2017universal}
S.-M. Moosavi-Dezfooli, A.~Fawzi, O.~Fawzi, and P.~Frossard.
\newblock Universal adversarial perturbations.
\newblock In {\em Proceedings of the IEEE Conference on Computer Vision and
  Pattern Recognition}, pages 1765--1773, 2017.

\bibitem{moosavi2016deepfool}
S.-M. Moosavi-Dezfooli, A.~Fawzi, and P.~Frossard.
\newblock Deepfool: a simple and accurate method to fool deep neural networks.
\newblock In {\em Proceedings of the IEEE Conference on Computer Vision and
  Pattern Recognition}, pages 2574--2582, 2016.

\bibitem{narwaria2012svd}
M.~Narwaria and W.~Lin.
\newblock Svd-based quality metric for image and video using machine learning.
\newblock {\em IEEE Transactions on Systems, Man, and Cybernetics, Part B
  (Cybernetics)}, 42(2):347--364, 2012.

\bibitem{netzer2011reading}
Y.~Netzer, T.~Wang, A.~Coates, A.~Bissacco, B.~Wu, and A.~Y. Ng.
\newblock Reading digits in natural images with unsupervised feature learning.
\newblock In {\em NIPS workshop on deep learning and unsupervised feature
  learning}, volume 2011, page~5, 2011.

\bibitem{Novikov2015}
A.~Novikov, D.~Podoprikhin, A.~Osokin, and D.~P. Vetrov.
\newblock Tensorizing neural networks.
\newblock In {\em Advances in Neural Information Processing Systems 28}, pages
  442--450. 2015.

\bibitem{oseledets2010tt}
I.~Oseledets and E.~Tyrtyshnikov.
\newblock Tt-cross approximation for multidimensional arrays.
\newblock {\em Linear Algebra and its Applications}, 432(1):70--88, 2010.

\bibitem{oseledets2011}
I.~V. Oseledets.
\newblock Tensor-train decomposition.
\newblock {\em SIAM Journal on Scientific Computing}, 33(5):2295--2317, 2011.

\bibitem{Papalexakis2016}
E.~E. Papalexakis, C.~Faloutsos, and N.~D. Sidiropoulos.
\newblock {Tensors for Data Mining and Data Fusion}.
\newblock {\em ACM Transactions on Intelligent Systems and Technology},
  8(2):1--44, 2016.

\bibitem{Papernot2016}
N.~Papernot, P.~McDaniel, A.~Sinha, and M.~Wellman.
\newblock Towards the science of security and privacy in machine learning.
\newblock {\em arXiv preprint arXiv:1611.03814}, 2016.

\bibitem{phan2010}
A.~H. Phan and A.~Cichocki.
\newblock Tensor decompositions for feature extraction and classification of
  high dimensional datasets.
\newblock {\em Nonlinear theory and its applications, IEICE}, 1(1):37--68,
  2010.

\bibitem{rudin1992nonlinear}
L.~I. Rudin, S.~Osher, and E.~Fatemi.
\newblock Nonlinear total variation based noise removal algorithms.
\newblock {\em Physica D: nonlinear phenomena}, 60(1-4):259--268, 1992.

\bibitem{russakovsky2015imagenet}
O.~Russakovsky, J.~Deng, H.~Su, J.~Krause, S.~Satheesh, S.~Ma, Z.~Huang,
  A.~Karpathy, A.~Khosla, M.~Bernstein, et~al.
\newblock Imagenet large scale visual recognition challenge.
\newblock {\em International Journal of Computer Vision}, 115(3):211--252,
  2015.

\bibitem{Sidiropoulos2016}
N.~D. Sidiropoulos, L.~D. Lathauwer, X.~Fu, K.~Huang, E.~E. Papalexakis, and
  C.~Faloutsos.
\newblock Tensor decomposition for signal processing and machine learning.
\newblock {\em IEEE Transactions on Signal Processing}, 65(13):3551--3582, July
  2017.

\bibitem{springenberg2014striving}
J.~T. Springenberg, A.~Dosovitskiy, T.~Brox, and M.~Riedmiller.
\newblock Striving for simplicity: The all convolutional net.
\newblock {\em arXiv preprint arXiv:1412.6806}, 2014.

\bibitem{stewart1990stochastic}
G.~W. Stewart.
\newblock Stochastic perturbation theory.
\newblock {\em SIAM review}, 32(4):579--610, 1990.

\bibitem{stewart1998perturbation}
G.~W. Stewart.
\newblock Perturbation theory for the singular value decomposition.
\newblock {\em In SVD and Signal Processing, II: Algorithms, Analysis and
  Applications}, 1991.

\bibitem{Sze2017}
V.~Sze, Y.-H. Chen, T.-J. Yang, and J.~S. Emer.
\newblock Efficient processing of deep neural networks: A tutorial and survey.
\newblock {\em Proceedings of the IEEE}, 105(12):2295--2329, 2017.

\bibitem{szegedy2016rethinking}
C.~Szegedy, V.~Vanhoucke, S.~Ioffe, J.~Shlens, and Z.~Wojna.
\newblock Rethinking the inception architecture for computer vision.
\newblock In {\em Proceedings of the IEEE conference on computer vision and
  pattern recognition}, pages 2818--2826, 2016.

\bibitem{tai2015}
C.~Tai, T.~Xiao, Y.~Zhang, X.~Wang, et~al.
\newblock Convolutional neural networks with low-rank regularization.
\newblock {\em International Conference on Learning Representations}, 2015.

\bibitem{tjandra2017compressing}
A.~Tjandra, S.~Sakti, and S.~Nakamura.
\newblock Compressing recurrent neural network with tensor train.
\newblock In {\em Neural Networks (IJCNN), 2017 International Joint Conference
  on}, pages 4451--4458. IEEE, 2017.

\bibitem{van1993subspace}
A.-J. Van Der~Veen, E.~F. Deprettere, and A.~L. Swindlehurst.
\newblock Subspace-based signal analysis using singular value decomposition.
\newblock {\em Proceedings of the IEEE}, 81(9):1277--1308, 1993.

\bibitem{vasilescu2002multilinear}
M.~A.~O. Vasilescu and D.~Terzopoulos.
\newblock Multilinear analysis of image ensembles: Tensorfaces.
\newblock In {\em European Conference on Computer Vision}, pages 447--460.
  Springer, 2002.

\bibitem{vervliet2016tensorlab}
N.~Vervliet, O.~Debals, and L.~De~Lathauwer.
\newblock Tensorlab 3.0 - numerical optimization strategies for large-scale
  constrained and coupled matrix/tensor factorization.
\newblock In {\em Signals, Systems and Computers, 2016 50th Asilomar Conference
  on}, pages 1733--1738. IEEE, 2016.

\bibitem{wang2018big}
X.~Wang, L.~T. Yang, H.~Liu, and M.~J. Deen.
\newblock A big data-as-a-service framework: State-of-the-art and perspectives.
\newblock {\em IEEE Transactions on Big Data}, 4(3):325--340, 2018.

\bibitem{wang2016}
Y.~Wang and A.~Anandkumar.
\newblock Online and differentially-private tensor decomposition.
\newblock In {\em Advances in Neural Information Processing Systems}, pages
  3531--3539, 2016.

\bibitem{xiao2018spatially}
C.~Xiao, J.-Y. Zhu, B.~Li, W.~He, M.~Liu, and D.~Song.
\newblock Spatially transformed adversarial examples.
\newblock {\em ICLR}, 2018.

\bibitem{xu2017feature}
W.~Xu, D.~Evans, and Y.~Qi.
\newblock Feature squeezing: Detecting adversarial examples in deep neural
  networks.
\newblock {\em In The Network and Distributed System Security Symposium
  (NDSS)}, 2017.

\bibitem{yang2015}
L.~T. Yang, L.~Kuang, J.~Chen, F.~Hao, and C.~Luo.
\newblock A holistic approach to distributed dimensionality reduction of big
  data.
\newblock {\em IEEE Transactions on Cloud Computing}, 2015.

\bibitem{yang2016}
Y.~Yang and T.~Hospedales.
\newblock Deep multi-task representation learning: A tensor factorisation
  approach.
\newblock {\em ICLR}, 2016.

\bibitem{yu2017}
S.~Yu, M.~Liu, W.~Dou, X.~Liu, and S.~Zhou.
\newblock Networking for big data: A survey.
\newblock {\em IEEE Communications Surveys \& Tutorials}, 19(1):531--549, 2017.

\bibitem{yunpeng2017}
C.~Yunpeng, J.~Xiaojie, K.~Bingyi, F.~Jiashi, and Y.~Shuicheng.
\newblock Sharing residual units through collective tensor factorization in
  deep neural networks.
\newblock {\em IJCAI}, 2017.

\bibitem{zhang2017improved}
Q.~Zhang, L.~T. Yang, Z.~Chen, and P.~Li.
\newblock An improved deep computation model based on canonical polyadic
  decomposition.
\newblock {\em IEEE Transactions on Systems, Man, and Cybernetics: Systems},
  2017.

\bibitem{zhang2018tensor}
Q.~Zhang, L.~T. Yang, Z.~Chen, and P.~Li.
\newblock A tensor-train deep computation model for industry informatics big
  data feature learning.
\newblock {\em IEEE Transactions on Industrial Informatics}, 2018.

\bibitem{zhang2017tucker}
Q.~Zhang, L.~T. Yang, X.~Liu, Z.~Chen, and P.~Li.
\newblock A tucker deep computation model for mobile multimedia feature
  learning.
\newblock {\em ACM Transactions on Multimedia Computing, Communications, and
  Applications (TOMM)}, 13(3s):39, 2017.

\bibitem{zhu2010automatic}
X.~Zhu and P.~Milanfar.
\newblock Automatic parameter selection for denoising algorithms using a
  no-reference measure of image content.
\newblock {\em IEEE transactions on image processing}, 19(12):3116--3132, 2010.

\end{thebibliography}
}

\end{document}